\documentclass{article}
\usepackage{ssba2005}
\usepackage{graphicx}
\usepackage{amssymb}
\usepackage{amsmath}
\usepackage{algorithmic}
\usepackage{algorithm}
\begin{document}

\title{Efficient Robust Mean Value Calculation \\ of 1D Features}
\author{Erik Jonsson and Michael Felsberg \\
        Computer Vision Laboratory \\
        Department of Electrical Enginering, Link\"{o}ping University, Sweden \\
        erijo@isy.liu.se, mfe@isy.liu.se}
\maketitle

\begin{abstract}
A robust mean value is often a good alternative to the standard mean value when dealing
with data containing many outliers. An efficient method for samples of one-dimensional
features and the truncated quadratic error norm is presented and compared to the method
of channel averaging (soft histograms).
\end{abstract}

\section{Introduction}

In a lot of applications in image processing we are faced with data containing lots of
outliers. One example is denoising and edge-preserving smoothing of low-level image
features, but the outlier problem also occurs in high-level operations like object
recognition and stereo vision. A wide range of robust techniques for different
applications have been presented, where RANSAC \cite{hartley} and the Hough transform
\cite{sonka} are two classical examples.

In this paper, we focus on the particular problem of calculating a mean value which is
robust against outliers. An efficient method for the special case of one-dimensional
features is presented and compared to the \emph{channel averaging} \cite{scia2003}
approach.

\section{Problem Formulation}
\label{sec:prob} 
\begin{figure}
  \center
  \includegraphics[width=0.2\textwidth]{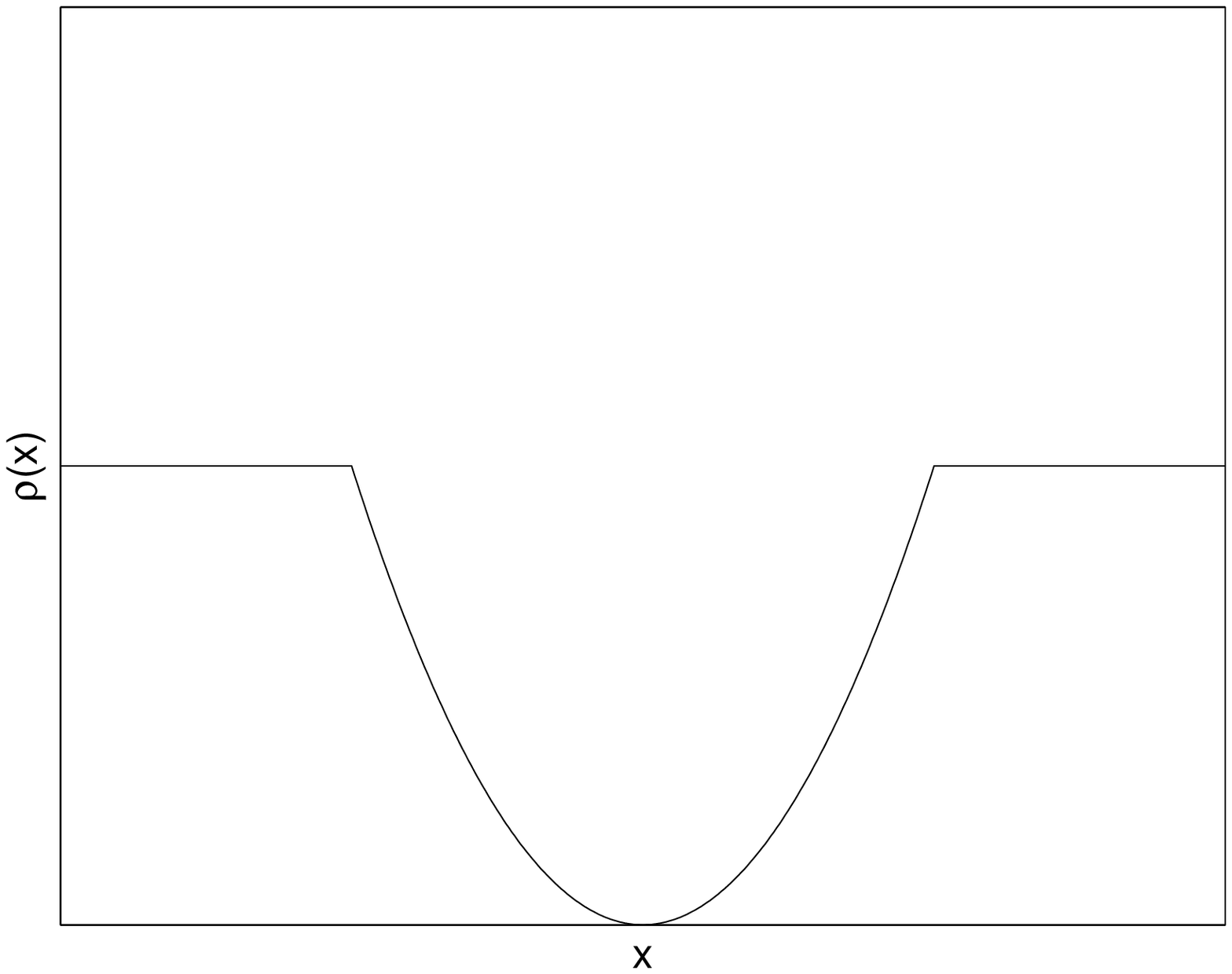}
  \includegraphics[width=0.2\textwidth]{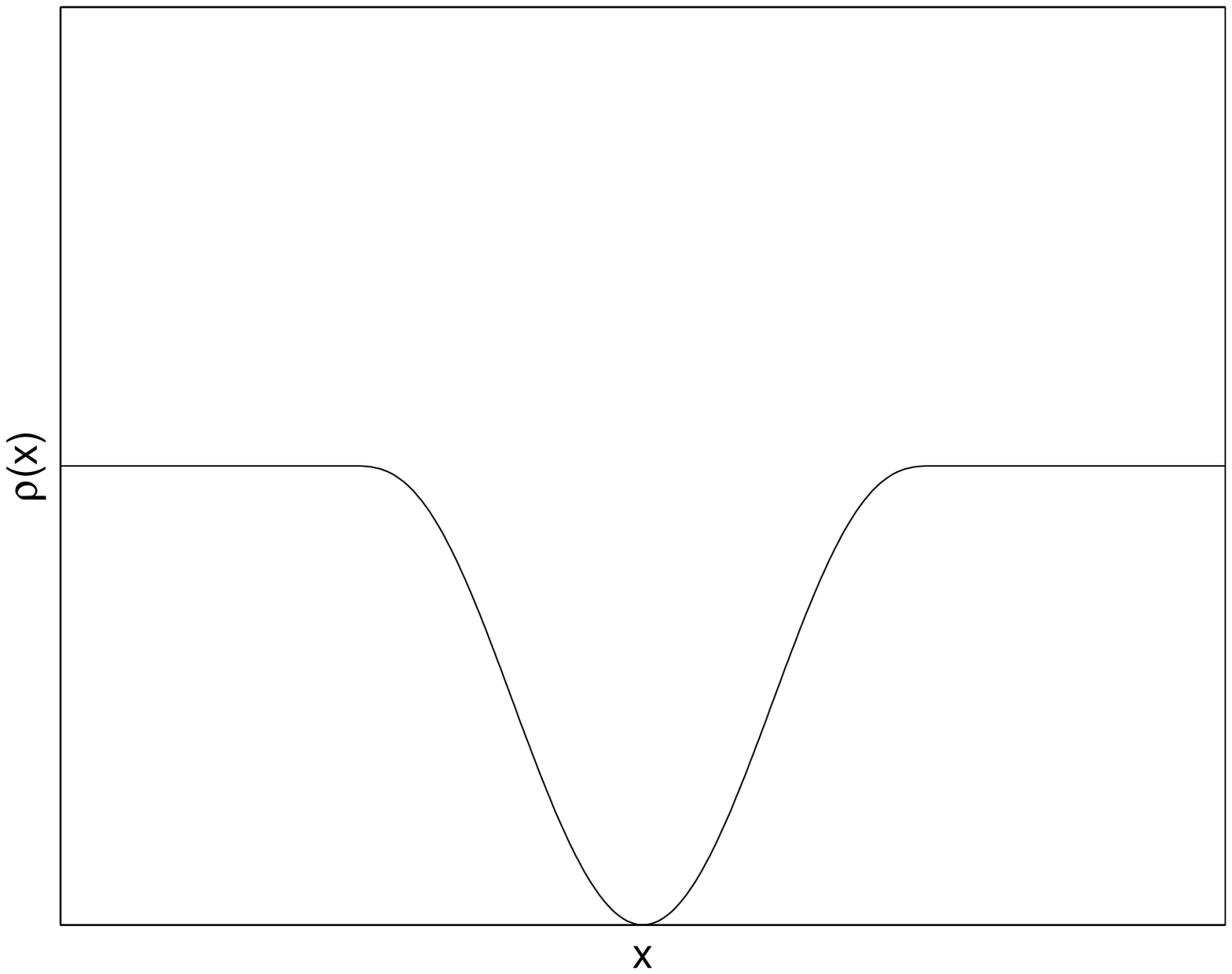}
  \caption{Error norms: Truncated quadratic (left), Tukey's biweight (right)}\label{fig:pf1}
\end{figure}

\begin{figure}
  \center
  \includegraphics[width=0.3\textwidth]{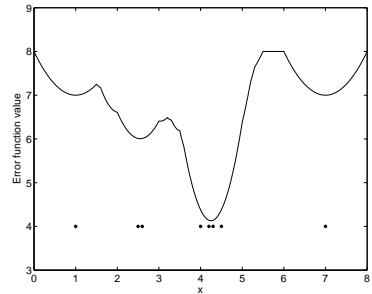}
  \caption{A simple 1D data set together with the error function generated using the truncated
  quadratic error norm with cutoff distance 1}\label{fig:pf2}
\end{figure}

Given a sample set $\mathbf{X} = [\mathbf{x}^{(1)} \ldots
\mathbf{x}^{(n)}]$, we seek to minimize an error function given by
\begin{equation}
  \label{eq:pf1}
   \mathcal{E}(\mathbf{x}) = \sum_{k=1}^n \rho(\| \mathbf{x}^{(k)} - \mathbf{x} \|)
\end{equation}
If we let $\rho$ be a quadratic function, the minimizing $\mathbf{x}$ is the standard
mean value. To achieve the desired robustness against outliers, $\rho$ should be a
function that saturates for large argument values. Such functions are called \emph{robust
error norms}. Some popular choices are the \emph{truncated quadratic} and \emph{Tukey's
biweight} shown in figure \ref{fig:pf1}. A simple 1D data set together with its error
function is shown in figure \ref{fig:pf2}. The $\mathbf{x}$ which minimizes
\eqref{eq:pf1} belongs to a general class of estimators called \emph{M-estimators}
\cite{Winkler02}, and will in this text be referred to as the \emph{robust mean value}.

\section{Previous Work}
\label{sec:prev}

Finding the robust mean is a non-convex optimization problem, and a unique global minimum
is not guaranteed. The problem is related to clustering, and the well-known \emph{mean
shift} iteration has been shown to converge to a local minimum of a robust error function
\cite{cheng95}.

% \subsection{Mean Shift}
% Finding the robust mean is a non-convex optimization problem, and a unique global optimum
% is not guaranteed. One of the most popular methods is the \emph{mean shift} iteration.
% The algorithm in its simplest form is outlined as follows: Pick a random starting point
% sample $\mu_0$. Let $\mu_1$ be the mean value of the samples within a window $\mu_0 \pm
% c$, and keep iterating until $\mu_k = \mu_{k+1}$. In \cite{cheng95}, it is shown that the
% mean shift converges to a local minimum of the error function (\ref{eq:pf1}).

\begin{figure}[t]
  \center
  \includegraphics[width=0.4\textwidth, height=0.2\textwidth]{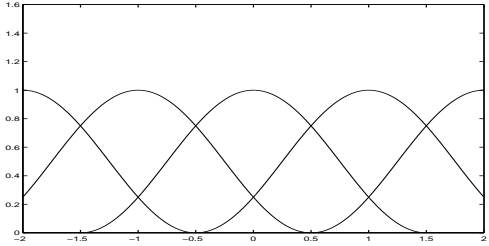}\\
  \caption{Example of channel kernel functions located at the integers}\label{fig:channel1}
\end{figure}

Another approach is to use the channel representation (soft histograms) \cite{f04a,
scia2003, fg04, Scharr03}. Each sample $\mathbf{x}$ can be encoded into a channel vector
$\mathbf{c}$ by the nonlinear transformation
\begin{equation}
  \label{eq:channel1}
  \mathbf{c} = [K(\|\mathbf{x}-\xi_1\|), \ldots, K(\|\mathbf{x}-\xi_m\|)]
\end{equation}
where $K$ is a localized kernel function and $\xi_k$ the \emph{channel centers},
typically located uniformly and such that the kernels overlap (fig \ref{fig:channel1}).
By averaging the channel representations of the samples, we get something which resembles
a histogram, but with overlapping and ``smooth'' bins. Depending on the choice of kernel,
the representation can be decoded to obtain an approximate robust mean. The distance
between neighboring channels corresponds to the scale of the robust error norm.

\section{Efficient 1D Method
  \protect\footnote{This section has been slightly revised since the
  original SSBA paper, as it contained some minor errors.}} \label{sec:fast}
This section will cover the case where the $\mathbf{x}$'s are one-dimensional,
\emph{e.g.} intensities in an image, and the truncated quadratic error norm is used. In
this case, there is a very efficient method, which we have not discovered in the
literature. For clarity, we describe the case where all samples have equal weight, but
the extension to weighted samples is straightforward.

First, some notation. We assume that our data is sorted in ascending order and numbered
from $1 \ldots n$. Since the $\mathbf{x}$'s are one dimensional, we drop the vector
notation and write simply $x_k$. The error norm is truncated at $c$, and can be written
as
\begin{equation}
  \rho(x) = \min\{x^2, c^2\}
\end{equation}
The method works as follows: We keep track of indices $a,b$ and and a window $w = [a,b]$
of samples $[x_a, \ldots, x_b]$. The window $[a,b]$ is said to be

\begin{itemize}
  \item[-] \emph{feasible} if $|x_b - x_a| < 2c$
  \item[-] \emph{maximal} if the samples are contained in a continuous window of length $2 c$,
    i.e. if $[a, b]$ is feasible and $[a-1, b+1]$ is infeasible.
\end{itemize}
Now define for a window $w = [a,b]$
\begin{eqnarray}
  \mu_w & = & \frac{1}{b-a+1} \sum_{k=a}^b x_k \\
  n_o & = & (a-1) + (n-b) \quad \\
  q_w & = & \sum_{k=a}^b (\mu_w - x_k)^2 \\
  \hat{\mathcal{E}}_w & = & q_w + n_o c^2
\end{eqnarray}
Note that $n_o$ is the number of samples outside the window. Consider the global minimum
$x_0$ of the error function and the window $w$ of samples $x_k$ that fall within the
quadratic part of the error function centered around $x_0$, i.e. the samples $x_k$ such
that $|x_k - x_0| \le c$. Either this window is located close to the boundary ($a=1$ or
$b=n$) or constitutes a maximal window. In both cases, $x_0 = \mu_w$, and
$\hat{\mathcal{E}}_w = \mathcal{E}(\mu_w)$. This is not necessarily true for an arbitrary
window, e.g. if $\mu_w$ is located close to the window boundary. However, for an
arbitrary window $w$, we have
\begin{eqnarray}
  \hat{\mathcal{E}}_w & = & \sum_{k=a}^b (\mu_w - x_k)^2 + n_o c^2 \ge \\
    & \ge & \sum_{k=1}^n \min\{(\mu_w - x_k)^2, c^2\} \\
   & = & \sum_{k=1}^n \rho(\mu_w - x_k) = \mathcal{E}(\mu_w)
\end{eqnarray}

The strategy is now to enumerate all maximal and boundary windows, evaluate
$\hat{\mathcal{E}}_w$ for each and take the minimum, which is guaranteed to be the global
minimum of $\mathcal{E}$. Note that it does not matter if some non-maximal windows are
included, since we always have $\hat{\mathcal{E}}_w \ge \mathcal{E}(\mu_w)$.

The following iteration does the job: Assume that we have a feasible window $[a, b]$, not
necessarily maximal. If $[a, b+1]$ is feasible, take this as the new window. Otherwise,
$[a, b]$ was the largest maximal window starting at $a$, and we should go on looking for
maximal windows starting at $a+1$. Take $[a+1, b]$ as the first candidate, then keep
increasing $b$ until the window becomes infeasible, etc. If proper initialization and
termination of the loop is provided, this iteration will generate all maximal and
boundary windows.

The last point to make is that we do not need to recompute $q_w$ from scratch as the
window size is changed. Similar to the treatment of mean values and variances in
statistics, we get by expanding the quadratic expression
\begin{eqnarray}
q_w & = & \sum_{k=a}^b (\mu_w - x_k)^2 =  \nonumber \\
    & = & \sum_{k=a}^b x_k^2 - (b-a+1) \mu_w^2 = \nonumber \\
    & = & S_2 - (b-a+1)^{-1} S_1^2
\end{eqnarray}
where we have defined
\begin{eqnarray}
  S_1 & = & \sum_{k=a}^b x_k = (b-a + 1) \mu_w \\
  S_2 & = & \sum_{k=a}^b x_k^2 \\
\end{eqnarray}
$S_1$ and $S_2$ can easily be updated in constant time as the window size is increased or
decreased, giving the whole algorithm complexity $O(n)$. The algorithm is summarized as
follows: \footnote{The check $a \le b$ is required to avoid zero division if $a$ was
increased beyond $b$ in the previous iteration.}

\begin{algorithm}
\caption{Fast 1D robust mean calculation} \label{alg:fast1}
\begin{algorithmic}
  \STATE Initialize $a \gets 1$, $b \gets 1$, $S_1 \gets x_1$, $S_2 \gets x_1^2$
  \WHILE{$a \le n$}
    \IF{$a \le b$}
      \STATE Calculate candidate $\hat{\mathcal{E}}_w$ and $\mu_w$:
      \STATE $\mu_w \gets (b-a+1)^{-1} S_1$
      \STATE $\hat{\mathcal{E}}_w \gets S_2 - \mu_w S_1 + n_o c^2$
      \STATE If $\hat{\mathcal{E}}_w$ is the smallest so far, store $\hat{\mathcal{E}}_w$, $\mu_w$.
    \ENDIF
    \IF{$b < n$ and $|x_{b+1} - x_a| < 2c$}
      \STATE $b \leftarrow b + 1$
      \STATE $S_1 \leftarrow S_1 + x_{b}$
      \STATE $S_2 \leftarrow S_2 + x_{b}^2$
    \ELSE
      \STATE $S_1 \leftarrow S_1 - x_{a}$
      \STATE $S_2 \leftarrow S_2 - x_{a}^2$
      \STATE $a \leftarrow a + 1$
    \ENDIF
  \ENDWHILE
  \STATE The $\mu_w$ corresponding to the smallest $\hat{\mathcal{E}}_w$ is now the robust mean.
\end{algorithmic}
\end{algorithm}

Note that it is straightforward to introduce a weight $w$ for each sample, such that a
weighted mean value is produced. We should then let $n_0$ be the total weight of the
samples outside the window, $\mu_w$ the weighted mean value of the window $w$, $S_1$ and
$S_2$ weighted sums etc.

\section{Properties of the Robust \\ Mean Value}

\begin{figure}[t]
  \center
  \includegraphics[width=0.4\textwidth]{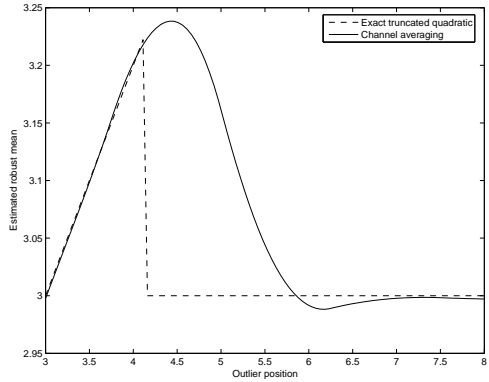}
  \caption{The influence of an outlier on the mean value}\label{fig:comp1}
\end{figure}

In this section, some properties of the robust mean values generated by the truncated
quadratic method and the channel averaging will be examined. In figure \ref{fig:comp1},
we show the robust mean of a sample set consisting of some values (inliers) with mean
value $3.0$ and an outlier at varying positions. As the outlier moves sufficiently far
away from the inliers, it is completely rejected, and when it is close to $3.0$, it is
treated as an inlier. As expected, the truncated quadratic method makes a hard decision
about whether the outlier should be included or not, whereas the channel averaging
implicitly assumes a smoother error norm.

\begin{figure}[t]
  \center
  \includegraphics[width=0.4\textwidth]{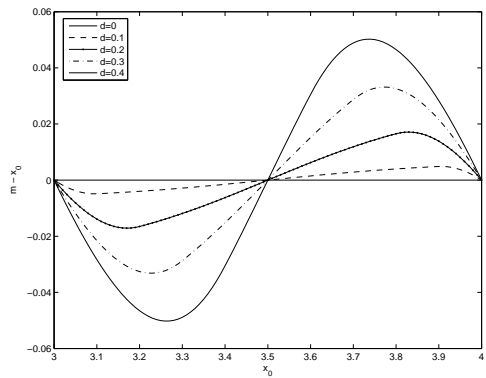}
  \caption{The grid effect}\label{fig:comp2}
\end{figure}

Another effect is that the channel averaging overcompensates for the outlier at some
positions (around $x=6.0$ in the plot). Also, the exact behavior of the method can vary
at different absolute positions due to the \emph{grid effect} illustrated in figure
\ref{fig:comp2}. We calculated the robust mean of two samples $x_1, x_2$, symmetrically
placed around some point $x_0$ with $|x_1-x_0| = |x_2-x_1| = d$. The channels were placed
with unit distance, and the displacement of the estimated mean $m$ compared to the
desired value $x_0$ is shown for varying $x_0$'s in the range between two neighboring
channel centers. The figure shows that the method makes some (small) systematic errors
depending on the position relative to the channel grid. No such grid effect occurs using
the method from section \ref{sec:fast}.

\begin{figure}[t]
  \center
  \includegraphics[width=0.4\textwidth]{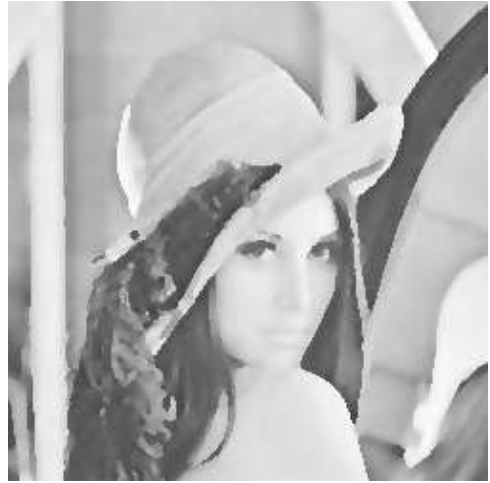}
  \caption{Lenna, robustly smoothed with the truncated quadratic method}\label{fig:lenna}
\end{figure}

When the robust mean algorithm is applied on sliding spatial windows of an image, we get
an edge-preserving image smoothing method. In figure \ref{fig:lenna}, we show the 256x256
Lenna image smoothed with the truncated quadratic method using a spatial window of 5 x 5
and $c = 0.1$ in the intensity domain, where intensities are in the range $[0,1]$. The
pixels are weighted with a Gaussian function.

\section{Discussion}

We have shown an efficient way to calculate the robust mean value for the special case of
one-dimensional features and the truncated quadratic error. The advantage of this method
is that it is simple, exact and global. The disadvantage is of course its limitation to
one-dimensional feature spaces.

One example of data for which the method could be applied is image features like
intensity or orientation. If the number of samples is high, e.g. in robust smoothing of a
high resolution image volume, the method might be suitable. If a convolution-like
operation is to be performed, the overhead of sorting the samples could be reduced
significantly, since the data is already partially sorted when moving to a new spatial
window, leading to an efficient edge-preserving smoothing algorithm.

\section*{Acknowledgment}
This work has been supported by EC Grant IST-2003-004176 COSPAL.

\bibliography{references}

\end{document}